\documentclass[a4paper]{cas-dc}
\usepackage[numbers]{natbib}
\usepackage{stfloats}
\usepackage{xurl}
\usepackage{float}
\usepackage[shortlabels]{enumitem}
\usepackage{rotating}
\usepackage{times}
\usepackage{makecell}
\usepackage{array}
\usepackage{multirow}
\newcolumntype{T}{>{\fontfamily{ptm}\selectfont}c}
\usepackage{caption}
\usepackage[ruled,vlined]{algorithm2e}
\def\tsc#1{\csdef{#1}{\textsc{\lowercase{#1}}\xspace}}
\tsc{WGM}
\tsc{QE}

\ExplSyntaxOn
\bool_gset_true:N \g_stm_nologo_bool
\ExplSyntaxOff

\begin{document}
\let\WriteBookmarks\relax
\def\floatpagepagefraction{1}
\def\textpagefraction{.001}

\shorttitle{~}
\shortauthors{Li et al.}  
\title [mode = title]{Multi-population Diversity-guided Genetic Algorithm for Feature Selection in Network Intrusion Detection}  

\author[1]{Chunzhen Li}
\ead{lcz309@stu.gdou.edu.cn}

\affiliation[1]{organization={School of Electronic and Information Engineering, Guangdong Ocean University},
            city={Zhanjiang},
            postcode={524088}, 
            state={Guangdong},
            country={China}}

\affiliation[2]{organization={School of Mathematics and Computer, Guangdong Ocean University},
            city={Zhanjiang},
            postcode={524088}, 
            state={Guangdong},
            country={China}}

\affiliation[3]{organization={School of Automation Engineering, University of Electronic Science and Technology of China},
            city={Chengdu},
            postcode={611731}, 
            state={Sichuan},
            country={China}}

\author[2]{Yueyong Tang}
\ead{tyy221@stu.gdou.edu.cn}

\author[2]{Jianyu Lai}
\ead{gdouljy115@stu.gdou.edu.cn}

\author[2,3]{Chuantao Li$^{\star}$}
\ead{11672411dd@stu.gdou.edu.cn}

\author[2]{Sheng Li$^{\star}$}
\ead{lish_ls@gdou.edu.cn}

\nonumnote{$^{\star}$Corresponding author.}
\RenewDocumentCommand \printorcid { } { }

\begin{abstract}
    Network Intrusion Detection System is a critical means of ensuring cybersecurity. However, existing Genetic Algorithm-based feature selection methods face several limitations when dealing with high-dimensional redundant traffic features. For example, population diversity is difficult to maintain, and evolutionary operators lack guidance. To solve these problems, this study proposes the Multi-Population Diversity-Guided Genetic Algorithm (MPDGGA). First, we build a chained multi-population evolutionary structure. Second, we introduce a diversity-guided operator based on information gain ratio. Experiments on NSL-KDD, UNSW-NB15, and 9 UCI datasets show that the proposed model significantly outperforms four other advanced multi-population feature selection models. Across the 11 datasets, it attains the highest accuracy on 10 datasets and at least 2.26\% of the features were selected.
\end{abstract}


\maketitle

\section{Introduction}
\label{sec:introduction} 

In the contemporary digital era, the internet has permeated every facet of socioeconomic infrastructure. New technology is growing fast and helps industries and economies to grow and change. But at the same time, there are more and more cyberattacks and break-in. This makes it harder for service providers and users to keep their systems safe \cite{Marwa-p1, He-p2,Rejin-p3}. building a strong and effective Network Intrusion Detection System (NIDS) is very important. It helps keep the secure operation of networked devices and critical business continuity.

NIDS continuously monitors network traffic to identify anomalies or malicious activities and trigger alerts. Depending on the detection mechanisms, these systems are generally categorized into two paradigms: anomaly-based detection and signature-based detection. In contrast to anomaly-based methods, signature-based detection methods excel at matching known attack patterns to accurately identify malicious behaviors, leading to their widespread adoption in practical scenarios \cite{Zoppi-p4,Luo-p5}. However, signature-based NIDS also encounters performance bottlenecks attributed to high-dimensional feature redundancy.

Network traffic data is often characterized by high-dimensional redundancy and complex nonlinear interactions. This phenomenon not only escalates the computational overhead of model training but also induces overfitting, resulting in elevated false positive rates in real-world network environments \cite{Wang-p6,Gong-p7}.

To mitigate these challenges, feature selection has emerged as a pivotal step in enhancing the efficiency and performance of NIDS. Existing methodologies are primarily classified into three categories: filter methods \cite{Raeisi-p8}, embedded methods \cite{Rahamathulla-p9}, and wrapper methods \cite{Umer-p10}.

Filter methods operate independently of the subsequent learning model. They typically employ statistical metrics like correlation, variance, or information gain to quickly screen features that are related to the label. These methods are computationally efficient for screening, but they often miss the complex relationships between features.

Embedded methods incorporate feature selection directly into the model training process. Techniques like LASSO and Random Forests can optimize model performance while simultaneously selecting features. However, their efficacy is heavily contingent upon specific classifier structures. This makes them less useful for handling complex nonlinear combinations.

Wrapper methods evaluate feature subsets based on the predictive performance of a classifier. They can find the best set of features overall. Because of this, wrapper methods are widely used for feature selection in NIDS.

Genetic Algorithm (GA) represent one of the most prominent wrapper methods. Their basic evolutionary structure and numerous variants are utilized a lot in many areas \cite{Holland-p11,Li-p12}. In NIDS, traffic features are usually not separate from each other. Instead, they are connected in complex ways and form feature groups. In such scenarios, different feature groups can depend on each other in many ways. This means that the contribution of a single feature to a label is significantly influenced by other features \cite{Xue-p13}.

By leveraging evolutionary mechanisms such as crossover and mutation operators, GA can effectively capture the intricate relationships within traffic features. Not only that, GA have a inherent advantage of parallel search capabilities \cite{Deng-p14,Nssibi-p15}.

Despite these advantages, existing GA variants still face several critical limitations when applied to NIDS:

\begin{enumerate}[label=\arabic*)]
\item When using a single population structure in high-dimensional feature spaces, the chromosomes tend to decay in population diversity. Consequently, the model often becomes trapped in local optima, resulting in premature convergence.
\item There is no quick way to check candidate feature subsets. This means the model has to call slow classifiers many times to validate random solutions. This severely constrains evolutionary efficiency and exploration capabilities.
\item Crossover and mutation points are picked at random. This causes the evolutionary process to lack effective guidance. This makes it challenging to identify key features with high discrimination and low redundancy.
\end{enumerate}

To deal with these deficiencies, this study proposes a Multi-Population Diversity-Guided Genetic Algorithm (MPDGGA) for feature selection in NIDS. The primary contributions are summarized as follows:

\begin{enumerate}[label=\arabic*)]
\item We propose a chained multi-population structure. This enhances local exploration within subpopulations. It also improves global exploration by sharing the elite chromosome information between subpopulations.
\item We design a feature subset evaluation criterion based on the information gain ratio. This allows for quick checks of candidate feature subsets. 
\item We develop a diversity-guided operator for evolution. It directs crossover and mutation by identifying key feature point with high discrimination and low redundancy. This maintains population diversity while enhancing convergence efficiency.
\end{enumerate}

The remainder of this study is organized as follows: Section \ref{sec:S2} reviews the relevant literature on feature selection; Section \ref{sec:S3} details the proposed model design and algorithmic implementation; Section \ref{sec:S4} presents the experimental results and discussion; Section \ref{sec:S5} concludes the study and outlines directions for future research.

\section{Related Work}\label{sec:S2} 

\subsection{NIDS Based on Feature Selection Techniques}

When building a robust NIDS, feature selection constitutes a pivotal phase for augmenting model efficacy and generalization capacity. By eliminating irrelevant or redundant features, feature selection not only mitigates computational overheads but also enhances the model's discrimination and interpretability \cite{Zorarpaci-p16}. Current feature selection methods are mostly grouped into three main types: filter methods, embedded methods, and wrapper methods.

Filter methods work independently of subsequent learning algorithms, typically quantifying feature-label correlations through statistical metrics for selection \cite{Hassan-p17}. For example, Akhone et al. \cite{Ahakonye-p18} used chi-square tests with Modified Decision Trees to identify SCADA traffic. They built feature sets based on how important the features were in the statistics. Despite these methods are fast and take little computing power, they often miss the complex connections between traffic features. Also, picking features based on set statistical cutoffs usually needs manual adjustments.

Embedded methods incorporate feature selection into the classifier training process, achieving dimensionality reduction through regularization constraints or feature importance \cite{Taha-p19, Gou-p20}. For example, Naqqad et al. \cite{Naqqad-p21} built a lightweight classification model for smart grid monitoring. They used LightGBM's feature importance to cut down the size of PMU data. These methods find a balance between being fast and working well. However, results of embedded methods often depend a lot on the specific model, this makes them less able to work well in different or new situations.

Wrapper methods pick feature sets based on classifier predictive performance, employ heuristic search strategies to identify optimal subsets within the feature space \cite{Hagar-p22}. For example, Srivastava et al. \cite{Srivastava-p23} built a lightweight hybrid model for resource-constrained intrusion detection. They used Grey Wolf Optimization to extract critical traffic features. 

Despite the diversification of wrapper methods in recent years, existing approaches commonly suffer from limitations such as monotonous population structures and ineffective evolutionary operators. In the context of NIDS, these deficiencies often precipitate premature convergence to local optima and hinder the maintenance of population diversity.

\subsection{Population Structure Improvement Strategy for Feature Selection Techniques}

Conventional heuristic algorithms predominantly employ a single-population evolutionary structure. In this scenario, all chromosomes go through selection, crossover, and mutation in the same unified space. Due to the absence of differentiated search and cooperative mechanisms, this single-population structure has some limitations.

The most significant of these is the monotonous population structure is susceptible to insufficient exploration and premature convergence in complex high-dimensional spaces. To augment search efficiency and robustness, researchers have proposed different multi-population structural designs. These designs fall into two main types.

The first approach decomposes the search space into multiple sub-tasks, assigning distinct sub-populations to explore corresponding subspaces \cite{Silva-p24}. For example, Hou et al. \cite{Hou-p25} proposed the Correlation-Induced Cooperative Evolution-Particle Swarm Optimization (CICCPSO). Their method groups features into high, medium, and low correlation spaces based on correlation strength. It then builds special steps to remove redundancy features for each group. Zhang et al. \cite{Zhang-p26} introduced the Competitive Swarm Optimizer with Context Vector Enhancement Strategy (CE-CCSO). It mitigates the tendency of traditional structures to become entrapped in local optima by replacing chromosomes with poor fitness.

The limitations of the first approach are mainly reflected in two aspects. First, the criteria for sub-space division struggle to maintain universality across diverse datasets; Second, when the feature space is complex and local optima are unevenly distributed, decomposition strategies may result in uneven allocation of search resources.

The second approach partitions the entire population into multiple subpopulations, each addressing the same task. By sharing high-quality information from interactive elite chromosomes, these subpopulations achieve collaborative search. This helps GA explore the whole space better and keeps population diversity \cite{Zhong-p27}. For example, Li et al. \cite{Li-p28} proposed the Multi-Population Evolutionary Algorithm for Feature Selection (MPEA-FS), it constructs multiple feature pools by combining Fisher scores and inflection point detection, utilizing guiding vectors to reduce redundant features. Li et al. \cite{Li-p29} proposed the real-coded Multi-Population Dynamic Competitive Genetic Algorithm (MPDCGA) for feature selection. It fuses mRMR and cosine similarity for subpopulation initialization, designing dynamic competitive operators and adaptive similarity crossover operators to enhance local exploitation and global exploration.

Overall, the second approach demonstrates better search capabilities in feature selection tasks. However, certain approaches introduce relatively complex initialization and encoding strategies, and there remains insufficient exploration of high-quality information exchange and multi-population cooperative evolution.

\subsection{Improvement Strategies for Evolutionary Operators in Feature Selection Techniques} 

In multi-population models for feature selection, the design of evolutionary operators directly influences the model's exploration-exploitation balance. Also, its capability to escape local optima, and overall convergence efficiency.

For example, Ren et al. \cite{Ren-p30} proposed an adaptive Genetic Algorithm integrating population dispersion and chromosome fitness, dynamically adjusting crossover and mutation probabilities through a dual adaptive mechanism to enhance convergence efficiency while preserving diversity. Too et al. \cite{Too-p31} introduced the Fast Competitive Genetic Algorithm (FRGA), employing a winner-loser group division and a three-parent crossover mechanism, while mitigating premature convergence through dynamic adjustment of crossover frequency. 

These approaches have ameliorated population diversity and search efficiency to a certain extent. However, the selection of crossover and mutation points, along with the combination of gene fragments, remains predominantly based on stochastic strategies in existing research. This results in a lack of clear evolutionary direction and generates a plethora of low-quality offspring.

To mitigate this issue, Kordos et al. \cite{Kordos-p32} proposed the Multi-Point Multi-Crossover (MPMC) method, which dynamically increases the permutations of gene fragment combinations to enhance genomic diversity. Zhou et al. \cite{Zhou-p33} introduced the Correlation-Guided Genetic Algorithm (CGGA), which constructs evaluation metrics by quantifying feature-label correlations and feature-feature redundancy. This enables the pre-assessment of candidate crossover point quality, thereby improving the efficacy of operator selection to a certain degree.

\section{Methodology} 
\label{sec:S3} 

\subsection{Model Overview}

In the field of NIDS, traffic feature spaces are frequently characterized by high-dimensional redundancy and complex nonlinear interactions. While GA exhibit robust global search capabilities, conventional signal-population structures often prove inadequate in high dimensional, complex feature spaces. These structures coupled with a lack of effective guidance in the search process, frequently precipitates a decay in population diversity and premature convergence.

Existing multi-population methods for feature selection primarily concentrate on the construction of subpopulation structures and chromosome allocation strategies. However, enhancing local exploration within subpopulations, facilitating the exchange of high-quality information between subpopulations, and improving the quality of candidate subsets for crossover and mutation operators remain critical challenges in augmenting feature selection performance.

\begin{figure*}[t!]
    \centering
    \includegraphics[width=1.0\linewidth]{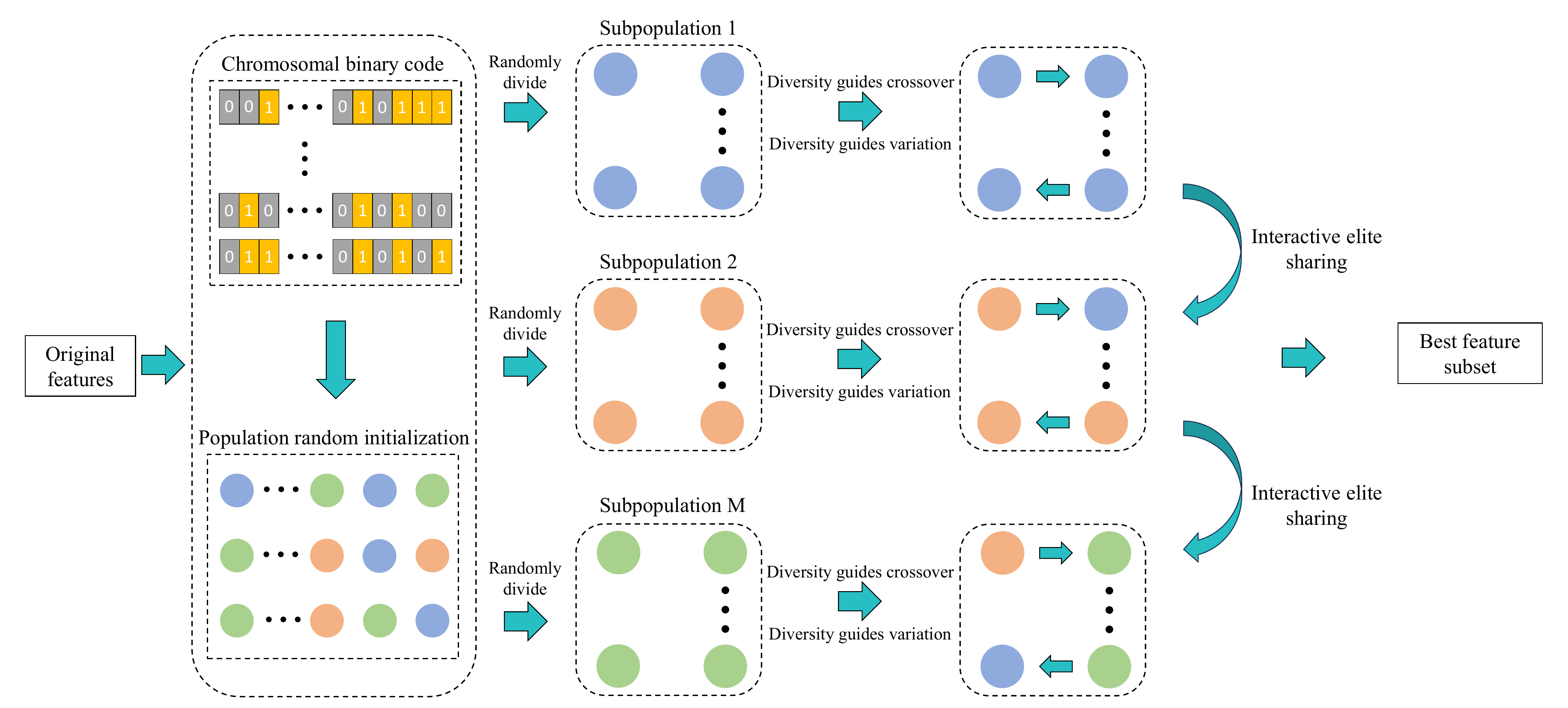}
    \caption{Overall flowchart of the model proposed in this study: Different colored circles represent chromosomes from different subpopulations.}
    \label{fig:Fig1} 
\end{figure*}

Consequently, this study proposes MPDGGA for feature selection in NIDS, with the comprehensive workflow illustrated in Figure\ref{fig:Fig1}.

Specifically, the proposed model first uses a binary encoding method to generate the initial population (refer to Section \ref{sec:S3.2}). Subsequently, the population is partitioned into multiple subpopulations and organized into a chained structure. Within this structure, crossover and mutation operations are executed within subpopulations, while elite interaction mechanisms facilitate the transfer of high-quality information across subpopulations (refer to Section \ref{sec:S3.3}). Furthermore, we introduce a subset evaluation criterion predicated on the information gain ratio. Based on this criterion, diversity-guided crossover and mutation operators are designed to enhance search efficiency and convergence quality while preserving population diversity (refer to Section \ref{sec:S3.4}). Finally, Section \ref{sec:S3.5} provides an analysis of the computational complexity to the MPDGGA.

\subsection{Chromosome Encoding and Population Initialization}
\label{sec:S3.2} 

\subsubsection{Binary Encoding Scheme} 

To formulate the feature selection task as a combinatorial optimization problem suitable for GA, we adopt a binary encoding scheme for chromosome representation. Each chromosome comprises a binary gene sequence of length $d$, where $d$ corresponds to the feature dimension. The mathematical relationship between chromosome $C$ and gene $g_k$ is defined as follows:

\begin{equation}
	C=(g_1,\cdots g_k,\cdots g_d ),\quad g_k\in\{0,1\},\quad 1\le k\le d,
\end{equation}

\noindent where $g_k=1$ signifies the inclusion of the $k$-th feature, whereas $g_k=0$ indicates its exclusion.

\subsubsection{Population Random Initialization} 

In feature selection, each chromosome represents a candidate feature subset. To guarantee maximal coverage of the feature space during the initialization phase, we employ a random initialization strategy. Given a population size of $N$, the population $P$ is mathematically represented as:

\begin{equation}
	P = 
    \begin{bmatrix} 
        g_{11} & g_{12} & \cdots & g_{1d} \\ 
        g_{21} & g_{22} & \cdots & g_{2d} \\ 
        \vdots & \vdots & \ddots & \vdots \\ 
        g_{N1} & g_{N2} & \cdots & g_{Nd} 
    \end{bmatrix}, 
    \quad g_{ij} \in \{0,1\},
\end{equation}
\noindent where $g_{ij}$ denotes the selection status of the $j$-th feature in the $i$-th chromosome.

\subsection{Multi-population Chain Structure} 
\label{sec:S3.3} 

Conventional GA predominantly utilize a single-population evolutionary structure. In this scenario, all chromosomes go through selection, crossover, and mutation in the same unified space. But in NIDS, the single-population structure often suffers from a dearth of effective mechanisms for information exchange and diversity preservation, rendering it susceptible to entrapment in local optima during the evolutionary process. 

To mitigate this limitation, we propose an enhanced multi-population chained structure. Within this structure, chromosomes within subpopulations are linked in a chain formation, undergoing independent crossover and mutation operations to bolster local exploration capabilities while sustaining population diversity. Concurrently, subpopulations are interlinked, enabling the exchange of high-quality information through the migration of elite chromosomes. This strategy significantly augments global exploration capabilities and diminishes the risk of subpopulations stagnating in local optima.

\subsubsection{Chromosomal Evolution within Subpopulations} 

\begin{figure*}[t!]
    \centering
    \includegraphics[width=1.0\linewidth]{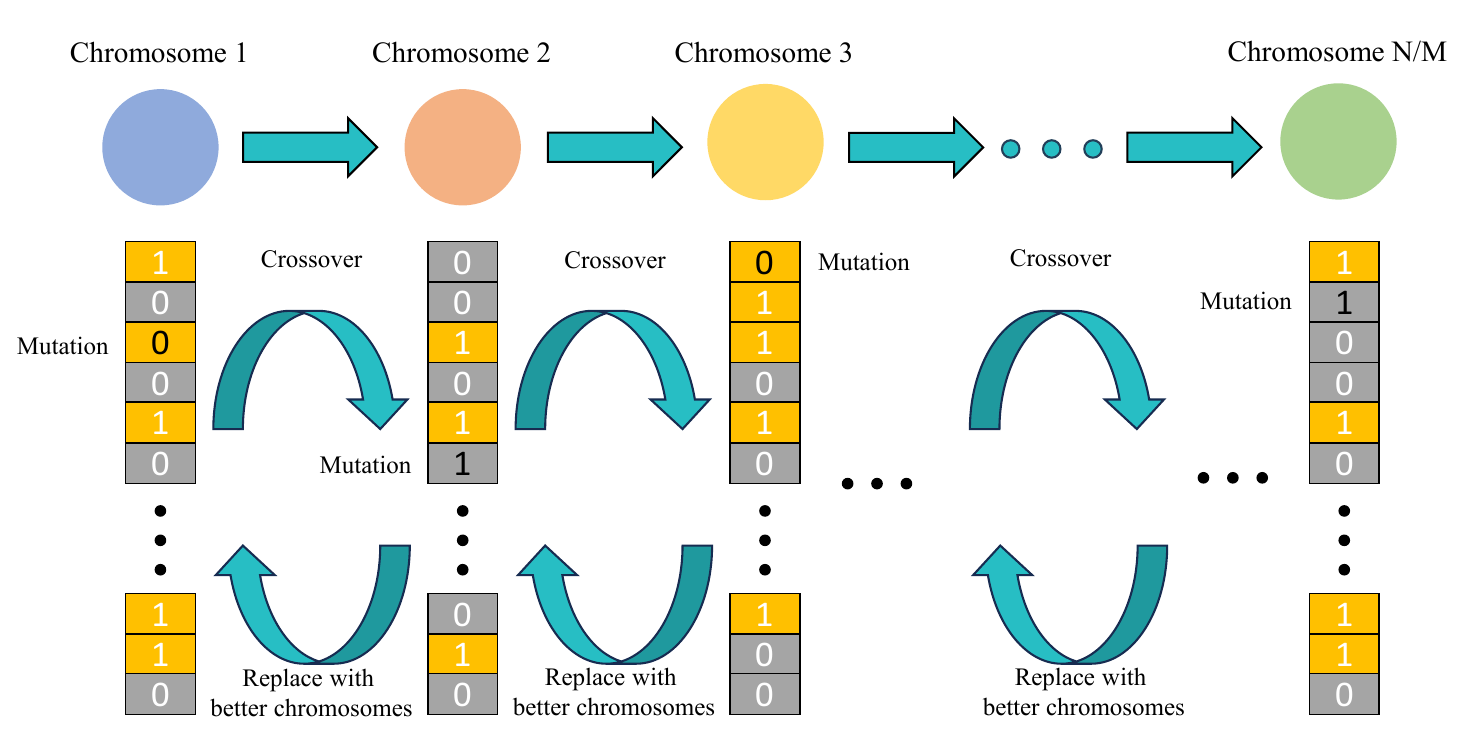}
    \caption{Chain structure of chromosomes within a subpopulation: Circles represent chromosomes, which are connected sequentially by chains. As shown by the arrows, the current chromosome crossover with the next chromosome, and then replaces the current chromosome based on the quality of the crossover result.}
    \label{fig:Fig2}
\end{figure*}

Following the initialization of population $P$, the population is randomly partitioned into $M$ subpopulations, each comprising $N/M$ chromosomes. As depicted in Figure \ref{fig:Fig2}, chromosomes within each subpopulation are sequentially linked to form a chain structure. The chromosome located at position $j$ within the $i$-th subpopulation is denoted as:

\begin{equation}
    \mathcal L_{i,j}, \quad i = 1,2, \dots, M, \quad j = 1,2, \dots,(N/M).
\end{equation}

In this intra-subpopulation chain structure, each chromosome sequentially engages in crossover operations, with its partner restricted to the subsequent chromosome in the chain. 

Using $\mathcal L_{i,j}$ as an illustrative example, a single-point crossover is performed with $\mathcal L_{i,j+1}$ to generate two offspring, $\mathcal L_{a}$ and $\mathcal L_{b}$. Subsequently, the fitness values of $\mathcal L_{i,j}$, $\mathcal L_{a}$, and $\mathcal L_{b}$ are evaluated, and the chromosome with the highest fitness replaces $\mathcal L_{i,j}$. 

Upon completion of the crossover phase, single-point gene mutations are executed on the chromosomes within the subpopulation, governed by a predefined mutation probability.

\subsubsection{Chromosomal Evolution among Subpopulations} 

While intra-subpopulation evolution emphasizes local optimization, the inter-subpopulation exchange of high-quality information by elite chromosome interaction. It is crucial for enhancing global exploration capabilities.

Following the completion of crossover and mutation within a subpopulation, the preceding subpopulation identifies $s$ elite chromosomes exhibiting the best fitness. These elites are then migrated to the subsequent subpopulation, replacing its $s$ least fit chromosomes. 

\begin{figure*}[t!]
    \centering
    \includegraphics[width=1.0\linewidth]{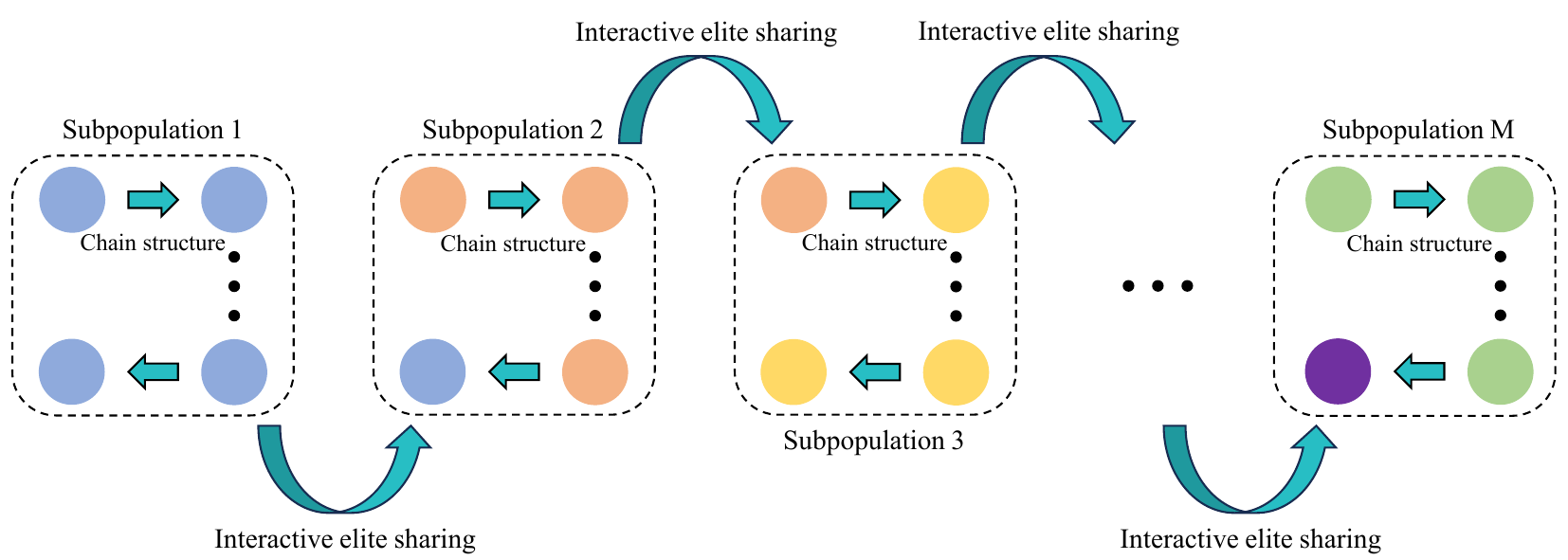}
    \caption{Chain structure of chromosomes among a subpopulation: The different colored circles appearing within the dashed boxes represent chromosomes that have been cross-substituted.}
    \label{fig:Fig3} 
\end{figure*}

As illustrated in Figure \ref{fig:Fig3}, elite chromosomes migrate sequentially along the chain structure, thereby facilitating the efficient transfer of high-quality information across subpopulations.

The fitness function serves as a metric for evaluating the quality of candidate solutions. The objective of feature selection is to minimize the size of the feature subset while maintaining optimal classification performance. Accordingly, we adopt the following fitness function $f(\mathcal L_{i,j})$:

\begin{equation}
    f(\mathcal L_{i,j})=(1-\alpha)\cdot \text{Acc}(\mathcal L_{i,j}) + \alpha \cdot \left(1-\frac{N_s}{N_f}\right),
    \label{eq:fitness}
\end{equation}

\noindent where $N_f$ represents the total number of original features, while $N_s$ denotes the number of features selected in the chromosome. The parameter $\alpha \in [0,1]$ serves as a trade-off coefficient, balancing classification performance and feature compression; in this study, $\alpha$ is set to $0.01$. $\text{Acc}(\mathcal L_{i,j})$ signifies the classification accuracy obtained using the feature subset selected by chromosome $\mathcal L_{i,j}$. In this study, the goal of the search is to find the feature subset that minimizes the fitness function.

\subsection{Diversity-guided Evolution Operators} 
\label{sec:S3.4} 

Conventional GA customarily employ single-point crossover and mutation operators. These random evolutionary operators suffer from a lack of directed search guidance, frequently generating a plethora of low-quality offspring. In wrapper-based methods, these random offspring necessitate evaluation by the classifier, resulting in substantial computational inefficiency. 

To cope with this issue, this study introduces a lightweight subset evaluation criterion. Prior to invoking the computationally expensive classifier, a rapid pre-evaluation of candidate feature subsets is conducted using statistical metrics. This strategy enables the prioritization of high-potential crossover points and mutation sites.

In contrast to filtering metrics such as correlation coefficients and symmetric uncertainty, the information gain ratio imposes penalties on features with highly discrete values that offer minimal contribution to category discrimination. Consequently, we construct a subset evaluation criterion predicated on the information gain ratio. 

Building upon this foundation, diversity-guided crossover and mutation operators are engineered to drive subpopulation evolution, thereby enhancing convergence quality while sustaining population diversity.

\subsubsection{Subset Evaluation Criteria Based on Information Gain Ratio} 
\label{sec:S3.4.1} 

Crossover or mutation operation yields a candidate feature subset. To avoid calling the costly classifiers too often to check the quality of these candidate solutions, we devise a subset evaluation rule based on the information gain ratio. For a given candidate feature subset $S$, its evaluation metric is defined as:

\begin{equation}
    J(S) = \frac{N_s \cdot \bar{R}_{FC}}{\sqrt{N_s + N_s(N_s-1)\bar{R}_{FF}}},
    \label{eq:score_criterion}
\end{equation}

\noindent where $N_s$ denotes the cardinality of the feature subset $S$. $\bar{R}_{FC}$ represents the average feature-class correlation, whereas $\bar{R}_{FF}$ denotes the average feature-feature redundancy. This rule helps us quickly measure how good a candidate feature subset is.

The average feature-class correlation $\bar{R}_{FC}$ quantifies the strength of the association between features within the subset and the target attack classes. It is calculated as follows:

\begin{equation}
    \bar{R}_{FC} = \frac{1}{N_s} \sum_{F_i \in S} \frac{IG(C, F_i)}{H(F_i)}.
\end{equation}

The average feature-feature redundancy $\bar{R}_{FF}$ measures the degree of redundancy among features within the set. In NIDS, if two traffic features exhibit high similarity, the simultaneous selection of both may not yield performance gains, even if they are associated with attack label. To foster the generation of diverse feature combinations during the evolutionary process, we define $\bar{R}_{FF}$ as:

\begin{equation}
    \bar{R}_{FF} =  \frac{1}{N_s(N_s-1)} \sum_{F_i \in S} \sum_{F_j \in S, j \neq i} \frac{IG(F_j, F_i)}{H(F_i)},
\end{equation}

\noindent where $IG(\cdot)$ denotes information gain, and $H(\cdot)$ represents information entropy.

\subsubsection{Diversity-guided Crossover Operator} 
\label{sec:S3.4.2} 

Single-point crossover generates offspring by exchanging gene segments between two parent chromosomes. Traditional single-point crossover randomly selects a crossover point $k \in [1, d-1]$ within the gene sequence and swaps the segments of parent chromosomes $\mathcal L_a$ and $\mathcal L_b$ beyond position $k$.

However, due to the intricate interactions among traffic features in NIDS, stably generating feature subsets that significantly enhance discrimination through random crossovers and mutations remains a formidable challenge. This often leads to diminished search efficiency and an elevated risk of premature convergence.

Therefore, leveraging the subset evaluation criterion, we propose a diversity-guided crossover operator. Given parent chromosomes $\mathcal L_a$ and $\mathcal L_b$, the operator iterates through all possible crossover positions $k \in \{1, 2, \dots, d-1\}$, generating two candidate offspring at each $k$:

\begin{equation}
    \begin{cases}
        \mathcal L_{a,k} = [\mathcal L_a(1 \dots k), \mathcal L_b(k+1 \dots d)], \\
        \mathcal L_{b,k} = [\mathcal L_b(1 \dots k), \mathcal L_a(k+1 \dots d)],
    \end{cases}
\end{equation}

\noindent subsequently, the quality of each candidate offspring pair, $J(\mathcal L_{a,k})$ and $J(\mathcal L_{b,k})$, is computed using the subset evaluation criterion. The optimal crossover point $k^*$ that maximizes offspring quality is then determined:

\begin{equation}
    k^* = \arg\max_{k}\left(J(\mathcal L_{a,k}), J(\mathcal L_{b,k})\right).
    \label{eq:crossover_k}
\end{equation}

Finally, the crossover operation is executed using $k^*$ to generate the optimized offspring.

\subsubsection{Diversity-guided Mutation Operator} 
\label{sec:S3.4.3} 

In binary encoding, single-point mutation works by flipping a gene. This happens with a chance set by the mutation probability $p_m$. Analogously, when mutation is triggered, the diversity-guided mutation operator goes through each gene spot $r$ from $1$ to $d$. For each spot, it creates candidate chromosomes $\mathcal L_{i,j}^{(r)}$ by flipping the bit at that spot. Then it figures out the evaluation scores $J(\mathcal L_{i,j}^{(r)})$ for these candidates. After that, it picks the best flipping spot $r^*$ like this:

\begin{equation}
    r^* = \arg\max_{r} J(\mathcal L_{i,j}^{(r)}).
    \label{eq:mutation_j}
\end{equation}

Finally, $\mathcal L_{i,j}$ is updated to $\mathcal L_{i,j}^{(r^*)}$. The pseudocode implementation of the proposed MPDGGA model is presented in Algorithm \ref{alg:MPDGGA}.

\begin{algorithm}
    \caption{Pseudocode implementation of the model proposed in this study.}
    \label{alg:MPDGGA}
    \small
    \SetAlgoLined
    \SetKwInOut{Input}{Input}
    \SetKwInOut{Output}{Output}

    \Input{
        Feature set $F = \{F_1, \dots, F_d\}$;\\ Iterations $T$;\\ Population size $N$;\\ Sub-populations $M$;\\ Number of elite chromosomes $s$; \\
        Mutation probability $p_m$
    }
    \Output{Optimal feature subset $S^*$}

    Initialize population $P = \{C_1, \dots, C_N\}$ with random binary encoding\;
    Pre-calculate information gain ratio matrix $\Omega$\;
    Divide $P$ into $M$ sub-populations $P_1,\dots,P_M$ and build chain structure\;
    \For{$t = 1$ \KwTo $T$}{
        \For{ $P_i,\ i \in \{1, \dots, M\}$ \textbf{Parallel}}{
            \For{$\mathcal{L}_{i,j} \in P_i$}{
                Select adjacent chromosomes in the chain as parents\;
                Select optimal crossover point $k^*$ for crossover; update $\mathcal{L}_{i,j}$ based on fitness\;
                \If{$\text{rand}(0,1) < p_m$}{
                    Select optimal mutation bit $r^*$ for mutation and update $\mathcal{L}_{i,j}$\;
                }
            }
        }
        \For{$i = 1$ \KwTo $M$}{
            Select top $s$ elites from $P_i$, migrate to $P_{(i \mod M) + 1}$ replacing its worst $s$ chromosomes\;
        }
    }
    \Return Global optimal feature subset $S^*$
\end{algorithm}

\subsection{Computational Complexity Analysis} 
\label{sec:S3.5} 

This section provides a theoretical analysis of the computational complexity inherent to the MPDGGA.

Let $d$ be the number of features, $n$ the number of samples, $T$ how many times the process repeats, and $N$ the size of the population. Before MPDGGA starts running, it needs to work out the information gain ratio matrix. This matrix captures how features relate to labels and also how features relate to each other. The result can be used again later in later rounds. This step takes $O(d^2 n)$ work.

In each round of evolution, MPDGGA runs diversity-guided crossover and mutation on $N$ chromosomes. For each one, they need to figure out a subset evaluation score. Because this evaluation uses average correlation and redundancy, it takes $O(d^2)$ work. So for one chromosome, the guided step takes $O(d^3)$ work. Also, the chosen chromosomes still have to be checked by the classifier. This check takes an amount of work called $O(f)$.

In summary, across $T$ iterations, the total time complexity of MPDGGA can be expressed as:

\begin{equation}
    O(d^2 n) + O(T \cdot N \cdot (d^3 + f)).
\end{equation}

Although the diversity-guided operator augments the computational cost per operation, its pre-evaluation mechanism significantly reduces the number of low-quality chromosomes entering the classifier evaluation phase. This reduction in classifier invocations ultimately leads to superior overall convergence efficiency.

\section{Experiments and Discussion} 
\label{sec:S4} 

\subsection{Experimental Setup} 

The experimental was implemented using Python 3.12. Essential dependencies, including scikit-learn (v1.5.1), Pandas (v2.2.2), and NumPy (v1.26.4), were managed via the Anaconda distribution. All tests in the experiment were run on a system with Ubuntu 22.04. The machine had 32 vCPUs from an AMD EPYC 9654 96-Core Processor and 60GB of memory.

\subsection{Dataset}

To empirically validate the efficacy and generalization capability of the proposed model, comprehensive experiments were conducted on NSL-KDD \cite{Tavallaee-NSLKDD}, UNSW-NB15 \cite{Moustafa-UNSW-NB15} and 9 UCI datasets \cite{Dua-UCI}. These datasets encompass a diverse range of sample sizes, feature dimensions, and class distributions, thereby facilitating a robust evaluation of the model's applicability across varied scenarios.

To ensure rigorous evaluation, all datasets were randomly partitioned into training, validation, and test sets following an 8:1:1 ratio. Table \ref{tab:all_datasets} details the statistical characteristics of each dataset.

\begin{table*}[htbp]
\centering
\caption{Statistical overview of the 11 datasets used in this study.}
\label{tab:all_datasets}
\fontfamily{ptm}\selectfont
\begin{tabular}{cccc}
\toprule
\textbf{Dataset} & \textbf{Number of Samples} & \textbf{Number of Features} & \textbf{Number of Classes} \\ 
\midrule
NSL-KDD                     & 148398    & 41    & 4  \\
UNSW-NB15                & 250982    & 42    & 6  \\
Arrhythmia                  & 452       & 279   & 7  \\
Darwin                      & 174       & 451   & 2  \\
Hill Valley                 & 606       & 101   & 2  \\
LSVT Voice Rehabilitation   & 126       & 310   & 2  \\
Parkinson                   & 756       & 754   & 2  \\
SPECTF                      & 267       & 44    & 2  \\
Sonar                       & 208       & 60    & 2  \\
Soybean                     & 307       & 35    & 15 \\
Spambase                    & 4601      & 57    & 2  \\
\bottomrule
\end{tabular}
\end{table*}

\subsection{Evaluation Metrics} 
\label{sec:metrics} 

To provide a comprehensive assessment of model performance across multi-class classification tasks, we utilize the following evaluation metrics:

\begin{enumerate}[label=(\arabic*)]
    \item \textbf{Accuracy}: Denotes the ratio of correctly classified samples to the total number of samples:
    \[
    \text{Accuracy} = \frac{\sum_{i=1}^{C} TP_i}{N},
    \]
    \item \textbf{Precision}: For category $i$, signifies the proportion of samples correctly identified as $i$ relative to all samples predicted as $i$:
    \[
    \text{Precision}_i = \frac{\text{TP}_i}{\text{TP}_i + \text{FP}_i},
    \]
    \item \textbf{Recall}: For class $i$, measures the fraction of actual class $i$ samples that are correctly identified:
    \[
    \text{Recall}_i = \frac{\text{TP}_i}{\text{TP}_i + \text{FN}_i},
    \]
    \item \textbf{F1 Score}: For class $i$, represents the harmonic mean of precision and recall, offering a balanced view of performance:
    \[
    F1_i = \frac{2 \cdot \text{Precision}_i \cdot \text{Recall}_i}{\text{Precision}_i + \text{Recall}_i},
    \]
\end{enumerate}

\noindent where $TP_i$, $TN_i$, $FP_i$, and $FN_i$ correspond to the counts of true positives, true negatives, false positives, and false negatives for category $i$, respectively. $C$ represents the total number of classes, and $N$ indicates the total sample size.

\subsection{Comparison Models} 

To benchmark the performance of MPDGGA, we selected a suite of representative multi-population and diversity-based evolutionary feature selection methods from recent literature for comparative analysis:

\begin{enumerate}[label=(\arabic*)]
    \item CGGA\cite{Zhou-p33}: Utilizes correlation information to steer genetic search, effectively reducing low-quality chromosomes and bolstering evolutionary efficiency.
    \item CE-CCSO\cite{Zhang-p26}: Employs a context vector enhancement strategy to alleviate the issue of coevolutionary stagnation in local optima.
    \item MPDCGA\cite{Li-p29}: Integrates real-coded, dynamic competition, and adaptive crossover mechanisms to augment diversity and circumvent premature convergence.
    \item MPEA-FS\cite{Li-p28}: A decomposable multi-population evolutionary model that harmonizes feature dimensionality with classification performance through collaborative subproblem resolution.
\end{enumerate}

To ensure a fair and rigorous comparison, all models were configured with a uniform iteration count of 30 and a population size of 30. The K-Nearest Neighbors algorithm, with $k=5$, served as the classifier. Detailed parameter settings are provided in Table \ref{tab:parasetting}.

\begin{table*}[ht]
\centering
\caption{Parameter settings for models tested in this study.}
\fontfamily{ptm}\selectfont
\label{tab:parasetting}
\begin{tabular}{cccccc}
\toprule
\textbf{Algorithm} & \textbf{Parameter} & \textbf{Value} & \textbf{Description} \\ 
\midrule
\multirow{2}{*}{CGGA}       & $k_1$        & 40   & Number of crossover attempts \\
           & $k_2$        & 40   & Number of mutation attempts \\
\midrule
\multirow{3}{*}{CE-CCSO}     & $\phi$       & 0.1  & Social factor \\
           & $\alpha$     & 5    & Stagnation threshold \\
           & $\mu$        & 0.3  & Subpopulation enhancement ratio \\
\midrule
\multirow{2}{*}{MPDCGA}     & $M$          & 3    & Number of subpopulations \\
           & $s$          & 2    & Number of interactive elites chromosomes \\
\midrule
MPEA-FS     & $M$          & 3    & Number of subpopulations \\
\midrule
\multirow{2}{*}{MPDGGA}      & $M$          & 3    & Number of subpopulations \\
           & $s$          & 2    & Number of interacting elites \\
\bottomrule
\end{tabular}
\end{table*}

\subsection{Performance Comparison}

Table \ref{tab:performance_combined} presents a comprehensive comparative analysis of the performance of various models across 11 datasets. On the two NIDS datasets, MPDGGA attained superior detection performance, highlighting its proficiency in handling large-scale traffic features. Specifically, on the NSL-KDD dataset, MPDGGA achieved an accuracy of 99.4\%. Compared to the second-best model, although the accuracy was similar, MPDGGA reduced the feature selection ratio by 9.76\% to 19.51\%.

The larger-scale UNSW-NB15 dataset, MPDGGA achieved 83.9\% accuracy while retaining only 33.3\% of the features. This empirical evidence demonstrates that the proposed multi-population cooperative architecture and diversity-guided mechanism can more effectively identify pivotal features and suppress redundancy.

Across the 9 UCI datasets, MPDGGA secured the highest accuracy on 8 datasets, demonstrating particularly pronounced advantages on high-dimensional datasets such as Arrhythmia, Darwin, and Parkinson. Taking the Parkinson dataset as an example, MPDGGA achieved an accuracy of 85.5\%, whereas CGGA, CE-CCSO, MPDCGA, and MPEA-FS achieved 75\%, 77.6\%, 71.1\%, and 76.3\%, respectively. On relatively small-scale datasets like Sonar and Soybean, MPDGGA also yields a more compact feature subset while maintaining high accuracy.

In summary, MPDGGA significantly outperformed other benchmark models across the 11 datasets. This is primarily attributed to the chained multi-population structure, which facilitates the transfer of high-quality information between subpopulations, while the diversity-guided operator mitigates the generation of low-quality offspring. This approach enhances global exploration capabilities while preserving population diversity.

\subsection{Ablation Study} 

To elucidate the chromosome contributions of each component module in MPDGGA, we conducted comprehensive ablation experiments across 11 datasets.

These experiments sequentially examined the effects of multi-population structure(M\_pop), elite interaction mechanism(E\_iter), diversity crossover operator(D\_cro), and diversity mutation operator(D\_mut). The results are systematically presented in Table \ref{tab:ablation_study}.

On the UNSW-NB15 dataset, introducing a multi-population structure improved the accuracy from 83.3\% to 83.5\%. Simultaneously, the feature selection rate decreased from 33.3\% to 30.9\%. This indicates that the multi-population structure effectively reduces feature redundancy while expanding the search scope.

Subsequently, integrating the elite interaction mechanism further improved the accuracy to 83.7\%. However, the feature selection rate slightly rebounded to 33.3\%. This suggests that while elite interaction promotes information sharing and improves model performance, it may require more refined control to avoid introducing a small number of redundant features.

Finally, the integration of diversity-guided operators further improved the accuracy to 83.9\% while maintaining a feature selection rate of 33.3\%, highlighting the crucial role of the guidance mechanism in promoting global exploration and maintaining feature brevity.

\subsection{Parameter Sensitivity Analysis} 

To analyze the impact of MPDGGA parameters on model performance, parameter sensitivity experiments were conducted across 11 datasets, with a focus on the effects of subpopulation size $M$ and elite interaction frequency $s$ on classification accuracy and feature selection ratio. The default settings were established as $M=3$ and $s=2$. The experimental results are detailed in Tables \ref{tab:parameter_sensitivity_m} and \ref{tab:parameter_sensitivity_s}.

On the NSL-KDD dataset, setting $M=1$ achieved the highest accuracy of 99.4\% while also obtaining the lowest feature selection rate of 19.51\%. As $M$ increased to 3 and 5, accuracy decreased to 98.9\% and the feature selection rate increased to 24.39\% and 26.83\%, respectively. For elite interaction on NSL-KDD, increasing $s$ from 2 to 5 improved accuracy from 98.8\% to 99.1\%. At the same time, the feature selection rate remained at 21.95\% for $s=2$, $s=5$, and $s=7$, while $s=3$ caused a noticeable increase to 29.27\%. These results indicate that a moderate-to-high interaction frequency can improve predictive performance without sacrificing subset compactness.

On the UNSW-NB15 dataset, the parameter effects were more explicit in terms of the accuracy-compactness trade-off. For subpopulation size, increasing $M$ from 1 to 3 improved accuracy from 78.6\% to 83.8\%, but the feature selection rate increased from 26.19\% to 33.33\%. Further increasing $M$ to 5 or 6 reduced accuracy to 80.0\% while lowering the feature selection rate to 28.57\%, indicating that larger subpopulation numbers did not yield additional performance gains. For elite interaction, increasing $s$ from 2 to 7 raised accuracy from 79.2\% to 81.1\%, whereas the lowest feature selection rate of 26.19\% was obtained at $s=5$. This pattern suggests that $s=7$ favors predictive performance, while $s=5$ favors feature compactness.

\section{Conclusion} 
\label{sec:S5} 

In NIDS, to address diversity loss, premature convergence, and weak search guidance in GA-based feature selection, this study proposes MPDGGA. Experimental results on NSL-KDD, UNSW-NB15, and 9 UCI datasets show that MPDGGA consistently improves classification performance while maintaining more compact feature subsets.

Notwithstanding the robust performance exhibited by MPDGGA across a diverse array of datasets, certain limitations remain to be addressed. Specifically, in scenarios characterized by pronounced nonlinear feature dependencies, the subset evaluation criterion predicated on information gain ratio may exhibit deficiencies in capturing intricate interaction patterns, thereby potentially impeding the identification of optimal feature coalitions.

Future research endeavors will prioritize two distinct trajectories. Firstly, the investigation of advanced subset evaluation criteria that incorporate auxiliary heuristic mechanisms to bolster the characterization of feature interaction structures. Secondly, the extension of MPDGGA's applicability to resource-constrained environments, such as the IoT and industrial control systems, to rigorously assess its deployability under real-time and lightweight constraints.

\clearpage
\begin{table*}[htbp]
\centering
\small
\caption{Performance Comparison of the Proposed Model and Comparison Models on 11 Datasets.}
\fontfamily{ptm}\selectfont
\label{tab:performance_combined}
\begingroup
\setlength{\tabcolsep}{3pt}
\renewcommand{\arraystretch}{0.95}
\begin{tabular*}{\textwidth}{@{\extracolsep{\fill}}ccccccc@{}}
\toprule
\textbf{Dataset} & \textbf{Model} & \textbf{Accuracy} & \textbf{Precision} & \textbf{Recall} & \textbf{F1-Score} & \textbf{FeatureRatio} \\ 
\midrule
\multirow{5}{*}{NSL-KDD} & Ours & \textbf{0.994} & \textbf{0.994} & \textbf{0.994} & \textbf{0.994} & \textbf{19.51\%} \\
                         & CGGA & 0.991 & 0.990 & 0.991 & 0.990 & 36.59\% \\
                         & CE-CCSO & 0.993 & 0.993 & 0.993 & 0.993 & 29.27\% \\
                         & MPDCGA & 0.993 & 0.993 & 0.993 & 0.993 & 39.02\% \\
                         & MPEA-FS & 0.993 & 0.993 & 0.993 & 0.993 & 24.39\% \\ \midrule
\multirow{5}{*}{UNSW-NB15} & Ours & \textbf{0.839} & \textbf{0.841} & \textbf{0.839} & \textbf{0.839} & 33.33\% \\
                           & CGGA & 0.833 & 0.835 & 0.833 & 0.834 & 45.24\% \\
                           & CE-CCSO & 0.830 & 0.835 & 0.830 & 0.832 & \textbf{28.57\%} \\
                           & MPDCGA & 0.830 & 0.831 & 0.830 & 0.830 & 33.33\% \\
                           & MPEA-FS & 0.692 & 0.680 & 0.692 & 0.681 & 33.33\% \\ \midrule
\multirow{5}{*}{Arrhythmia} & Ours & \textbf{0.571} & \textbf{0.595} & \textbf{0.571} & \textbf{0.543} & \textbf{8.96\%} \\
                            & CGGA & 0.429 & 0.238 & 0.429 & 0.286 & 30.11\% \\
                            & CE-CCSO & 0.429 & 0.257 & 0.429 & 0.306 & 37.28\% \\
                            & MPDCGA & 0.429 & 0.238 & 0.429 & 0.286 & 42.65\% \\
                            & MPEA-FS & 0.429 & 0.257 & 0.429 & 0.306 & 37.63\% \\ \midrule
\multirow{5}{*}{Darwin} & Ours & 0.778 & 0.792 & 0.778 & 0.775 & \textbf{19.87\%} \\
                        & CGGA & 0.611 & 0.625 & 0.611 & 0.600 & 32.59\% \\
                        & CE-CCSO & 0.722 & 0.725 & 0.722 & 0.721 & 47.32\% \\
                        & MPDCGA & 0.611 & 0.612 & 0.611 & 0.610 & 45.31\% \\
                        & MPEA-FS & \textbf{0.833} & \textbf{0.837} & \textbf{0.833} & \textbf{0.833} & 45.76\% \\ \midrule
\multirow{5}{*}{Hill Valley} & Ours & \textbf{0.631} & \textbf{0.631} & \textbf{0.631} & \textbf{0.631} & 38.00\% \\
                             & CGGA & 0.582 & 0.582 & 0.582 & 0.580 & \textbf{37.00\%} \\
                             & CE-CCSO & 0.598 & 0.600 & 0.598 & 0.594 & 44.00\% \\
                             & MPDCGA & 0.623 & 0.625 & 0.623 & 0.620 & 48.00\% \\
                             & MPEA-FS & \textbf{0.631} & \textbf{0.631} & \textbf{0.631} & \textbf{0.631} & 39.00\% \\ \midrule
\multirow{5}{*}{LSVT Voice Rehabilitation} & Ours & \textbf{0.923} & \textbf{0.931} & \textbf{0.923} & \textbf{0.920} & \textbf{2.26\%} \\
                                           & CGGA & 0.615 & 0.462 & 0.615 & 0.527 & 36.45\% \\
                                           & CE-CCSO & 0.692 & 0.657 & 0.692 & 0.656 & 34.84\% \\
                                           & MPDCGA & 0.692 & 0.657 & 0.692 & 0.656 & 41.29\% \\
                                           & MPEA-FS & 0.538 & 0.441 & 0.538 & 0.485 & 34.52\% \\ \midrule
\multirow{5}{*}{Parkinson} & Ours & \textbf{0.855} & \textbf{0.858} & \textbf{0.855} & \textbf{0.856} & \textbf{24.17\%} \\
                           & CGGA & 0.750 & 0.722 & 0.750 & 0.723 & 44.36\% \\
                           & CE-CCSO & 0.776 & 0.767 & 0.776 & 0.770 & 41.57\% \\
                           & MPDCGA & 0.711 & 0.683 & 0.711 & 0.692 & 48.21\% \\
                           & MPEA-FS & 0.763 & 0.750 & 0.763 & 0.754 & 37.45\% \\ \midrule
\multirow{5}{*}{SPECTF} & Ours & \textbf{0.852} & \textbf{0.852} & \textbf{0.852} & \textbf{0.852} & \textbf{27.27\%} \\
                        & CGGA & 0.815 & 0.805 & 0.815 & 0.809 & 40.91\% \\
                        & CE-CCSO & 0.741 & 0.756 & 0.741 & 0.748 & 31.82\% \\
                        & MPDCGA & 0.741 & 0.690 & 0.741 & 0.706 & \textbf{27.27\%} \\
                        & MPEA-FS & 0.667 & 0.645 & 0.667 & 0.655 & 34.09\% \\ \midrule
\multirow{5}{*}{Sonar} & Ours & \textbf{0.810} & \textbf{0.810} & \textbf{0.810} & \textbf{0.810} & \textbf{11.67\%} \\
                       & CGGA & 0.762 & 0.765 & 0.762 & 0.762 & 28.33\% \\
                       & CE-CCSO & 0.667 & 0.667 & 0.667 & 0.665 & 16.67\% \\
                       & MPDCGA & 0.714 & 0.714 & 0.714 & 0.714 & 36.67\% \\
                       & MPEA-FS & 0.714 & 0.714 & 0.714 & 0.714 & 13.33\% \\ \midrule
\multirow{5}{*}{Soybean} & Ours & \textbf{0.926} & \textbf{0.914} & \textbf{0.926} & \textbf{0.912} & \textbf{34.29\%} \\
                         & CGGA & 0.889 & 0.870 & 0.889 & 0.877 & 37.14\% \\
                         & CE-CCSO & 0.778 & 0.784 & 0.778 & 0.755 & 42.86\% \\
                         & MPDCGA & 0.852 & 0.852 & 0.852 & 0.840 & 45.71\% \\
                         & MPEA-FS & 0.815 & 0.811 & 0.815 & 0.802 & \textbf{34.29\%} \\ \midrule
\multirow{5}{*}{Spambase} & Ours & \textbf{0.915} & \textbf{0.916} & \textbf{0.915} & \textbf{0.916} & 45.61\% \\
                          & CGGA & 0.894 & 0.894 & 0.894 & 0.893 & \textbf{38.59\%} \\
                          & CE-CCSO & 0.902 & 0.902 & 0.902 & 0.902 & 49.12\% \\
                          & MPDCGA & 0.911 & 0.911 & 0.911 & 0.911 & 50.87\% \\
                          & MPEA-FS & 0.898 & 0.898 & 0.898 & 0.898 & 50.87\% \\
\bottomrule
\end{tabular*}
\endgroup
\end{table*}

\clearpage
\begin{table*}[htbp]
\centering
\caption{Ablation Analysis Results of the Proposed Model on 11 Datasets.}
\fontfamily{ptm}\selectfont
\label{tab:ablation_study}
\footnotesize
\begingroup
\setlength{\tabcolsep}{3.0pt}
\renewcommand{\arraystretch}{0.95}
\begin{tabular*}{\textwidth}{@{\extracolsep{\fill}}cccccccccc@{}}
\toprule
\textbf{Dataset} & \textbf{M\_pop} & \textbf{E\_iter} & \textbf{D\_cro} & \textbf{D\_mut} & \textbf{Accuracy} & \textbf{Precision} & \textbf{Recall} & \textbf{F1-Score} & \textbf{FeatureRatio} \\ \midrule
\multirow{6}{*}{NSL-KDD} & $\times$ & $\times$ & $\times$ & $\times$ & 0.986 & 0.986 & 0.986 & 0.985 & 17.10\% \\
 & $\checkmark$ & $\times$ & $\times$ & $\times$ & 0.987 & 0.987 & 0.987 & 0.986 & 31.70\% \\
 & $\checkmark$ & $\checkmark$ & $\times$ & $\times$ & 0.989 & 0.989 & 0.989 & 0.988 & 19.51\% \\
 & $\checkmark$ & $\checkmark$ & $\checkmark$ & $\times$ & 0.990 & 0.990 & 0.990 & 0.989 & 21.95\% \\
 & $\checkmark$ & $\checkmark$ & $\times$ & $\checkmark$ & 0.992 & 0.992 & 0.992 & 0.991 & 24.39\% \\
 & $\checkmark$ & $\checkmark$ & $\checkmark$ & $\checkmark$ & \textbf{0.994} & \textbf{0.994} & \textbf{0.994} & \textbf{0.994} & \textbf{19.51\%} \\
\midrule
\multirow{6}{*}{UNSW-NB15} & $\times$ & $\times$ & $\times$ & $\times$ & 0.833 & 0.833 & 0.833 & 0.832 & 33.33\% \\
 & $\checkmark$ & $\times$ & $\times$ & $\times$ & 0.835 & 0.835 & 0.835 & 0.834 & \textbf{30.95\%} \\
 & $\checkmark$ & $\checkmark$ & $\times$ & $\times$ & 0.837 & \textbf{0.841} & 0.837 & 0.838 & 33.33\% \\
 & $\checkmark$ & $\checkmark$ & $\checkmark$ & $\times$ & 0.838 & 0.838 & 0.838 & 0.837 & 33.33\% \\
 & $\checkmark$ & $\checkmark$ & $\times$ & $\checkmark$ & 0.835 & 0.837 & 0.835 & 0.836 & 38.10\% \\
 & $\checkmark$ & $\checkmark$ & $\checkmark$ & $\checkmark$ & \textbf{0.839} & \textbf{0.841} & \textbf{0.839} & \textbf{0.839} & 33.33\% \\
\midrule
\multirow{6}{*}{Arrhythmia} & $\times$ & $\times$ & $\times$ & $\times$ & 0.571 & 0.429 & 0.571 & 0.476 & 33.33\% \\
 & $\checkmark$ & $\times$ & $\times$ & $\times$ & 0.571 & 0.400 & 0.571 & 0.449 & 19.71\% \\
 & $\checkmark$ & $\checkmark$ & $\times$ & $\times$ & \textbf{0.714} & \textbf{0.714} & \textbf{0.714} & \textbf{0.667} & 16.13\% \\
 & $\checkmark$ & $\checkmark$ & $\checkmark$ & $\times$ & 0.571 & 0.543 & 0.571 & 0.497 & 9.32\% \\
 & $\checkmark$ & $\checkmark$ & $\times$ & $\checkmark$ & 0.429 & 0.238 & 0.429 & 0.286 & 11.83\% \\
 & $\checkmark$ & $\checkmark$ & $\checkmark$ & $\checkmark$ & 0.571 & 0.595 & 0.571 & 0.543 & \textbf{8.96\%} \\
\midrule
\multirow{6}{*}{Darwin} & $\times$ & $\times$ & $\times$ & $\times$ & 0.611 & 0.781 & 0.611 & 0.542 & 38.62\% \\
 & $\checkmark$ & $\times$ & $\times$ & $\times$ & 0.722 & 0.750 & 0.722 & 0.714 & 35.94\% \\
 & $\checkmark$ & $\checkmark$ & $\times$ & $\times$ & 0.722 & 0.750 & 0.722 & 0.714 & 26.79\% \\
 & $\checkmark$ & $\checkmark$ & $\checkmark$ & $\times$ & 0.611 & 0.625 & 0.611 & 0.600 & 21.21\% \\
 & $\checkmark$ & $\checkmark$ & $\times$ & $\checkmark$ & \textbf{0.833} & \textbf{0.837} & \textbf{0.833} & \textbf{0.833} & 27.90\% \\
 & $\checkmark$ & $\checkmark$ & $\checkmark$ & $\checkmark$ & 0.778 & 0.792 & 0.778 & 0.775 & \textbf{19.87\%} \\
\midrule
\multirow{6}{*}{Hill Valley} & $\times$ & $\times$ & $\times$ & $\times$ & 0.631 & 0.634 & 0.631 & 0.628 & 43.00\% \\
 & $\checkmark$ & $\times$ & $\times$ & $\times$ & 0.631 & 0.631 & 0.631 & 0.631 & 36.00\% \\
 & $\checkmark$ & $\checkmark$ & $\times$ & $\times$ & 0.631 & 0.632 & 0.631 & 0.630 & 41.00\% \\
 & $\checkmark$ & $\checkmark$ & $\checkmark$ & $\times$ & \textbf{0.664} & \textbf{0.665} & \textbf{0.664} & \textbf{0.663} & 34.00\% \\
 & $\checkmark$ & $\checkmark$ & $\times$ & $\checkmark$ & 0.639 & 0.642 & 0.639 & 0.636 & \textbf{33.00\%} \\
 & $\checkmark$ & $\checkmark$ & $\checkmark$ & $\checkmark$ & 0.631 & 0.631 & 0.631 & 0.631 & 38.00\% \\
\midrule
\multirow{6}{*}{\makecell[l]{LSVT Voice Rehabilitation}} & $\times$ & $\times$ & $\times$ & $\times$ & 0.692 & 0.657 & 0.692 & 0.656 & 24.84\% \\
 & $\checkmark$ & $\times$ & $\times$ & $\times$ & 0.692 & 0.657 & 0.692 & 0.656 & 13.23\% \\
 & $\checkmark$ & $\checkmark$ & $\times$ & $\times$ & 0.615 & 0.587 & 0.615 & 0.598 & 14.84\% \\
 & $\checkmark$ & $\checkmark$ & $\checkmark$ & $\times$ & 0.692 & 0.657 & 0.692 & 0.656 & 5.16\% \\
 & $\checkmark$ & $\checkmark$ & $\times$ & $\checkmark$ & 0.692 & 0.657 & 0.692 & 0.656 & 8.39\% \\
 & $\checkmark$ & $\checkmark$ & $\checkmark$ & $\checkmark$ & \textbf{0.923} & \textbf{0.931} & \textbf{0.923} & \textbf{0.920} & \textbf{2.26\%} \\
\midrule
\multirow{6}{*}{Parkinson} & $\times$ & $\times$ & $\times$ & $\times$ & 0.776 & 0.767 & 0.776 & 0.770 & 35.99\% \\
 & $\checkmark$ & $\times$ & $\times$ & $\times$ & 0.763 & 0.750 & 0.763 & 0.754 & 32.40\% \\
 & $\checkmark$ & $\checkmark$ & $\times$ & $\times$ & 0.763 & 0.750 & 0.763 & 0.754 & 27.89\% \\
 & $\checkmark$ & $\checkmark$ & $\checkmark$ & $\times$ & 0.776 & 0.767 & 0.776 & 0.770 & \textbf{19.79\%} \\
 & $\checkmark$ & $\checkmark$ & $\times$ & $\checkmark$ & 0.789 & 0.779 & 0.789 & 0.781 & 29.88\% \\
 & $\checkmark$ & $\checkmark$ & $\checkmark$ & $\checkmark$ & \textbf{0.855} & \textbf{0.858} & \textbf{0.855} & \textbf{0.856} & 24.17\% \\
\midrule
\multirow{6}{*}{SPECTF} & $\times$ & $\times$ & $\times$ & $\times$ & 0.815 & 0.796 & 0.815 & 0.790 & 50.00\% \\
 & $\checkmark$ & $\times$ & $\times$ & $\times$ & 0.741 & 0.690 & 0.741 & 0.706 & \textbf{18.18\%} \\
 & $\checkmark$ & $\checkmark$ & $\times$ & $\times$ & 0.815 & 0.796 & 0.815 & 0.790 & 27.27\% \\
 & $\checkmark$ & $\checkmark$ & $\checkmark$ & $\times$ & 0.741 & 0.690 & 0.741 & 0.706 & 31.82\% \\
 & $\checkmark$ & $\checkmark$ & $\times$ & $\checkmark$ & 0.667 & 0.645 & 0.667 & 0.655 & 45.45\% \\
 & $\checkmark$ & $\checkmark$ & $\checkmark$ & $\checkmark$ & \textbf{0.852} & \textbf{0.852} & \textbf{0.852} & \textbf{0.852} & 27.27\% \\
\midrule
\multirow{6}{*}{Sonar} & $\times$ & $\times$ & $\times$ & $\times$ & 0.762 & 0.782 & 0.762 & 0.755 & 18.33\% \\
 & $\checkmark$ & $\times$ & $\times$ & $\times$ & \textbf{0.810} & \textbf{0.820} & \textbf{0.810} & 0.807 & 11.67\% \\
 & $\checkmark$ & $\checkmark$ & $\times$ & $\times$ & 0.762 & 0.763 & 0.762 & 0.761 & \textbf{6.67\%} \\
 & $\checkmark$ & $\checkmark$ & $\checkmark$ & $\times$ & 0.524 & 0.524 & 0.524 & 0.524 & 8.33\% \\
 & $\checkmark$ & $\checkmark$ & $\times$ & $\checkmark$ & 0.762 & 0.782 & 0.762 & 0.755 & 16.67\% \\
 & $\checkmark$ & $\checkmark$ & $\checkmark$ & $\checkmark$ & \textbf{0.810} & 0.810 & \textbf{0.810} & \textbf{0.810} & 11.67\% \\
\midrule
\multirow{6}{*}{Soybean} & $\times$ & $\times$ & $\times$ & $\times$ & 0.852 & 0.835 & 0.852 & 0.824 & 31.43\% \\
 & $\checkmark$ & $\times$ & $\times$ & $\times$ & 0.889 & 0.895 & 0.889 & 0.875 & \textbf{25.71\%} \\
 & $\checkmark$ & $\checkmark$ & $\times$ & $\times$ & \textbf{0.926} & \textbf{0.926} & \textbf{0.926} & \textbf{0.926} & 28.57\% \\
 & $\checkmark$ & $\checkmark$ & $\checkmark$ & $\times$ & 0.889 & 0.896 & 0.889 & 0.876 & 34.29\% \\
 & $\checkmark$ & $\checkmark$ & $\times$ & $\checkmark$ & 0.815 & 0.833 & 0.815 & 0.802 & 28.57\% \\
 & $\checkmark$ & $\checkmark$ & $\checkmark$ & $\checkmark$ & \textbf{0.926} & 0.914 & \textbf{0.926} & 0.912 & 34.29\% \\
\midrule
\multirow{6}{*}{Spambase} & $\times$ & $\times$ & $\times$ & $\times$ & 0.905 & 0.904 & 0.905 & 0.904 & 52.63\% \\
 & $\checkmark$ & $\times$ & $\times$ & $\times$ & 0.896 & 0.896 & 0.896 & 0.896 & \textbf{42.11\%} \\
 & $\checkmark$ & $\checkmark$ & $\times$ & $\times$ & 0.900 & 0.900 & 0.900 & 0.900 & 45.61\% \\
 & $\checkmark$ & $\checkmark$ & $\checkmark$ & $\times$ & 0.911 & 0.912 & 0.911 & 0.910 & \textbf{42.11\%} \\
 & $\checkmark$ & $\checkmark$ & $\times$ & $\checkmark$ & 0.902 & 0.903 & 0.902 & 0.903 & \textbf{42.11\%} \\
 & $\checkmark$ & $\checkmark$ & $\checkmark$ & $\checkmark$ & \textbf{0.915} & \textbf{0.916} & \textbf{0.915} & \textbf{0.916} & 45.61\% \\
\bottomrule
\end{tabular*}
\endgroup
\end{table*}

\clearpage
\begin{table*}[htbp]
\centering
\small
\caption{Parameter Sensitivity Analysis of Subpopulations Number on 11 Datasets.}
\fontfamily{ptm}\selectfont
\label{tab:parameter_sensitivity_m}
\begingroup
\setlength{\tabcolsep}{3pt}
\renewcommand{\arraystretch}{1.0}
\begin{tabular*}{\textwidth}{@{\extracolsep{\fill}}cccccccc@{}}
\toprule
\textbf{Dataset} & \textbf{Parameter} & \textbf{Value} & \textbf{Accuracy} & \textbf{Precision} & \textbf{Recall} & \textbf{F1-Score} & \textbf{FeatureRatio} \\
\midrule
\multirow{4}{*}{NSL-KDD} & M & 1 & \textbf{0.994} & \textbf{0.994} & \textbf{0.994} & \textbf{0.994} & \textbf{19.51\%} \\
 & M & 3 & 0.989 & 0.988 & 0.989 & 0.988 & 24.39\% \\
 & M & 5 & 0.989 & 0.989 & 0.989 & 0.989 & 26.83\% \\
 & M & 6 & 0.990 & 0.990 & 0.990 & 0.990 & 21.95\% \\
\midrule
\multirow{4}{*}{UNSW-NB15} & M & 1 & 0.786 & 0.785 & 0.786 & 0.784 & \textbf{26.19\%} \\
 & M & 3 & \textbf{0.839} & \textbf{0.841} & \textbf{0.839} & \textbf{0.839} & 33.33\% \\
 & M & 5 & 0.800 & 0.797 & 0.800 & 0.798 & 28.57\% \\
 & M & 6 & 0.800 & 0.797 & 0.800 & 0.798 & 28.57\% \\
\midrule
\multirow{4}{*}{Arrhythmia} & M & 1 & 0.286 & 0.167 & 0.286 & 0.210 & 15.05\% \\
 & M & 3 & 0.429 & 0.429 & 0.429 & 0.429 & 10.39\% \\
 & M & 5 & \textbf{0.571} & \textbf{0.595} & \textbf{0.571} & \textbf{0.543} & 8.96\% \\
 & M & 6 & 0.429 & 0.524 & 0.429 & 0.448 & \textbf{7.53\%} \\
\midrule
\multirow{4}{*}{Darwin} & M & 1 & 0.500 & 0.500 & 0.500 & 0.498 & 25.72\% \\
 & M & 3 & \textbf{0.778} & \textbf{0.792} & \textbf{0.778} & \textbf{0.775} & 19.73\% \\
 & M & 5 & 0.722 & 0.725 & 0.722 & 0.721 & 20.62\% \\
 & M & 6 & 0.667 & 0.667 & 0.667 & 0.667 & \textbf{13.30\%} \\
\midrule
\multirow{4}{*}{Hill Valley} & M & 1 & 0.549 & 0.549 & 0.549 & 0.545 & 44.55\% \\
 & M & 3 & 0.598 & 0.600 & 0.598 & 0.594 & 42.57\% \\
 & M & 5 & 0.500 & 0.498 & 0.500 & 0.495 & 41.58\% \\
 & M & 6 & \textbf{0.631} & \textbf{0.631} & \textbf{0.631} & \textbf{0.631} & \textbf{37.62\%} \\
\midrule
\multirow{4}{*}{\makecell[l]{LSVT Voice Rehabilitation}} & M & 1 & 0.769 & 0.759 & 0.769 & 0.759 & 13.87\% \\
 & M & 3 & \textbf{0.923} & \textbf{0.931} & \textbf{0.923} & \textbf{0.920} & \textbf{2.26\%} \\
 & M & 5 & 0.692 & 0.692 & 0.692 & 0.692 & 3.87\% \\
 & M & 6 & 0.769 & 0.759 & 0.769 & 0.759 & \textbf{2.26\%} \\
\midrule
\multirow{4}{*}{Parkinson} & M & 1 & 0.737 & 0.729 & 0.737 & 0.732 & 31.30\% \\
 & M & 3 & 0.803 & 0.795 & 0.803 & 0.797 & 21.88\% \\
 & M & 5 & 0.724 & 0.728 & 0.724 & 0.726 & 25.73\% \\
 & M & 6 & \textbf{0.829} & \textbf{0.826} & \textbf{0.829} & \textbf{0.828} & \textbf{20.56\%} \\
\midrule
\multirow{4}{*}{SPECTF} & M & 1 & 0.704 & 0.664 & 0.704 & 0.681 & 25.00\% \\
 & M & 3 & 0.741 & 0.690 & 0.741 & 0.706 & 43.18\% \\
 & M & 5 & \textbf{0.815} & \textbf{0.850} & \textbf{0.815} & \textbf{0.759} & 52.27\% \\
 & M & 6 & 0.667 & 0.645 & 0.667 & 0.655 & \textbf{18.18\%} \\
\midrule
\multirow{4}{*}{Sonar} & M & 1 & 0.667 & 0.677 & 0.667 & 0.657 & 18.33\% \\
 & M & 3 & \textbf{0.714} & 0.746 & \textbf{0.714} & \textbf{0.701} & \textbf{10.00\%} \\
 & M & 5 & 0.667 & 0.677 & 0.667 & 0.657 & \textbf{10.00\%} \\
 & M & 6 & \textbf{0.714} & \textbf{0.815} & \textbf{0.714} & 0.684 & \textbf{10.00\%} \\
\midrule
\multirow{4}{*}{Soybean} & M & 1 & 0.815 & 0.815 & 0.815 & 0.810 & \textbf{28.57\%} \\
 & M & 3 & 0.741 & 0.716 & 0.741 & 0.720 & 31.43\% \\
 & M & 5 & \textbf{0.852} & \textbf{0.840} & \textbf{0.852} & \textbf{0.826} & 45.71\% \\
 & M & 6 & 0.815 & 0.796 & 0.815 & 0.796 & 31.43\% \\
\midrule
\multirow{4}{*}{Spambase} & M & 1 & 0.892 & 0.891 & 0.892 & 0.891 & 47.37\% \\
 & M & 3 & 0.894 & 0.894 & 0.894 & 0.894 & \textbf{38.60\%} \\
 & M & 5 & 0.887 & 0.889 & 0.887 & 0.888 & 49.12\% \\
 & M & 6 & \textbf{0.898} & \textbf{0.898} & \textbf{0.898} & \textbf{0.898} & 40.35\% \\
\bottomrule
\end{tabular*}
\endgroup
\end{table*}

\clearpage
\begin{table*}[htbp]
\centering
\small
\caption{Parameter Sensitivity Analysis of Interacting Elites Number on 11 Datasets.}
\fontfamily{ptm}\selectfont
\label{tab:parameter_sensitivity_s}
\begingroup
\setlength{\tabcolsep}{3pt}
\renewcommand{\arraystretch}{1.0}
\begin{tabular*}{\textwidth}{@{\extracolsep{\fill}}cccccccc@{}}
\toprule
\textbf{Dataset} & \textbf{Parameter} & \textbf{Value} & \textbf{Accuracy} & \textbf{Precision} & \textbf{Recall} & \textbf{F1-Score} & \textbf{FeatureRatio} \\
\midrule
\multirow{4}{*}{NSL-KDD} & S & 2 & 0.988 & 0.988 & 0.988 & 0.988 & \textbf{21.95\%} \\
 & S & 3 & 0.989 & 0.989 & 0.989 & 0.989 & 29.27\% \\
 & S & 5 & \textbf{0.991} & \textbf{0.990} & \textbf{0.991} & \textbf{0.990} & \textbf{21.95\%} \\
 & S & 7 & 0.990 & \textbf{0.990} & 0.990 & \textbf{0.990} & \textbf{21.95\%} \\
\midrule
\multirow{4}{*}{UNSW-NB15} & S & 2 & 0.792 & 0.790 & 0.792 & 0.790 & 35.71\% \\
 & S & 3 & 0.807 & 0.806 & 0.807 & 0.805 & 33.33\% \\
 & S & 5 & 0.807 & 0.800 & 0.807 & 0.801 & \textbf{26.19\%} \\
 & S & 7 & \textbf{0.811} & \textbf{0.811} & \textbf{0.811} & \textbf{0.810} & 30.95\% \\
\midrule
\multirow{4}{*}{Arrhythmia} & S & 2 & \textbf{0.429} & 0.310 & \textbf{0.429} & 0.352 & 12.54\% \\
 & S & 3 & 0.286 & 0.167 & 0.286 & 0.210 & 10.04\% \\
 & S & 5 & \textbf{0.429} & 0.357 & \textbf{0.429} & 0.381 & \textbf{6.81\%} \\
 & S & 7 & \textbf{0.429} & \textbf{0.429} & \textbf{0.429} & \textbf{0.429} & 12.54\% \\
\midrule
\multirow{4}{*}{Darwin} & S & 2 & \textbf{0.667} & 0.667 & \textbf{0.667} & \textbf{0.667} & \textbf{19.73\%} \\
 & S & 3 & \textbf{0.667} & \textbf{0.675} & \textbf{0.667} & 0.662 & 21.95\% \\
 & S & 5 & 0.556 & 0.558 & 0.556 & 0.550 & 22.39\% \\
 & S & 7 & \textbf{0.667} & 0.667 & \textbf{0.667} & \textbf{0.667} & \textbf{19.73\%} \\
\midrule
\multirow{4}{*}{Hill Valley} & S & 2 & \textbf{0.566} & \textbf{0.566} & \textbf{0.566} & 0.561 & \textbf{36.63\%} \\
 & S & 3 & \textbf{0.566} & \textbf{0.566} & \textbf{0.566} & \textbf{0.564} & \textbf{36.63\%} \\
 & S & 5 & 0.500 & 0.499 & 0.500 & 0.499 & 37.62\% \\
 & S & 7 & 0.533 & 0.532 & 0.533 & 0.531 & 39.60\% \\
\midrule
\multirow{4}{*}{\makecell[l]{LSVT Voice Rehabilitation}} & S & 2 & 0.615 & 0.587 & 0.615 & 0.598 & 5.16\% \\
 & S & 3 & 0.692 & 0.692 & 0.692 & 0.692 & \textbf{1.61\%} \\
 & S & 5 & \textbf{0.846} & \textbf{0.874} & \textbf{0.846} & 0.828 & \textbf{1.61\%} \\
 & S & 7 & \textbf{0.846} & 0.846 & \textbf{0.846} & \textbf{0.846} & 4.19\% \\
\midrule
\multirow{4}{*}{Parkinson} & S & 2 & \textbf{0.855} & \textbf{0.858} & \textbf{0.855} & \textbf{0.856} & 24.14\% \\
 & S & 3 & 0.776 & 0.767 & 0.776 & 0.770 & 23.74\% \\
 & S & 5 & 0.776 & 0.780 & 0.776 & 0.778 & \textbf{23.08\%} \\
 & S & 7 & 0.776 & 0.762 & 0.776 & 0.765 & 23.47\% \\
\midrule
\multirow{4}{*}{SPECTF} & S & 2 & 0.741 & 0.725 & 0.741 & 0.732 & \textbf{25.00\%} \\
 & S & 3 & 0.778 & 0.754 & 0.778 & 0.761 & \textbf{25.00\%} \\
 & S & 5 & \textbf{0.852} & \textbf{0.852} & \textbf{0.852} & \textbf{0.852} & 27.27\% \\
 & S & 7 & 0.741 & 0.725 & 0.741 & 0.732 & 27.27\% \\
\midrule
\multirow{4}{*}{Sonar} & S & 2 & 0.524 & 0.524 & 0.524 & 0.524 & 13.33\% \\
 & S & 3 & \textbf{0.810} & \textbf{0.810} & \textbf{0.810} & \textbf{0.810} & 11.67\% \\
 & S & 5 & 0.571 & 0.570 & 0.571 & 0.569 & \textbf{8.33\%} \\
 & S & 7 & 0.667 & 0.667 & 0.667 & 0.665 & 15.00\% \\
\midrule
\multirow{4}{*}{Soybean} & S & 2 & 0.852 & 0.852 & 0.852 & 0.846 & 37.14\% \\
 & S & 3 & 0.815 & 0.774 & 0.815 & 0.790 & 40.00\% \\
 & S & 5 & \textbf{0.926} & \textbf{0.914} & \textbf{0.926} & \textbf{0.912} & \textbf{34.29\%} \\
 & S & 7 & 0.741 & 0.716 & 0.741 & 0.720 & 40.00\% \\
\midrule
\multirow{4}{*}{Spambase} & S & 2 & \textbf{0.915} & \textbf{0.916} & \textbf{0.915} & \textbf{0.916} & 45.61\% \\
 & S & 3 & 0.889 & 0.891 & 0.889 & 0.890 & 43.86\% \\
 & S & 5 & 0.900 & 0.901 & 0.900 & 0.900 & 40.35\% \\
 & S & 7 & 0.905 & 0.906 & 0.905 & 0.905 & \textbf{36.84\%} \\
\bottomrule
\end{tabular*}
\endgroup
\end{table*}

\clearpage

\section*{Credit}

\textbf{Chunzhen Li}: Investigation, Project administration, Validation, Visualization. \textbf{Yueyong Tang}: Data curation, Resources.\textbf{Jianyu Lai}: Data curation, Software.\textbf{Chuantao Li}: Conceptualization, Methodology, Writing – original draft.\textbf{Sheng Li}: Funding acquisition, Supervision, Writing – review and editing.

\section*{Data availability}

The datasets used and analyzed in the current study are available from the corresponding authors upon reasonable request.

\section*{Acknowledgments}

We sincerely thank the anonymous reviewers for their time and great efforts in reviewing our paper, their valuable comments, suggestions, and advice helped us significantly improve the quality of our manuscript. This work is supported by the National College Students Innovation and Entrepreneurship Training Program (No. 202510566026) and Guangdong Ocean University Undergraduate Innovation Team Project (No. CXTD2023014).

\printcredits

\bibliographystyle{cas-model2-names}

\bibliography{thesis} 

\end{document}